\documentclass[letterpaper, 10 pt, conference]{ieeeconf}

\IEEEoverridecommandlockouts                              % This command is only needed if 
\usepackage{graphicx} % for pdf, bitmapped graphics files
\usepackage{amsmath} % assumes amsmath package installed
\usepackage{multirow}
\usepackage{subcaption}

\title{\LARGE \bf
Self-supervised classification of dynamic obstacles using the temporal information provided by videos
}

\author{Sid Ali Hamideche$^{1}$, Florent Chiaroni$^{1, 2}$ and Mohamed-Cherif Rahal$^{1}$% <-this % stops a space
\thanks{$^{1}$ VEDECOM Institute, Department of delegated driving (VEH08), Perception team, 23 bis Allee des Marronniers, 78000 Versailles, France
        {\tt\small \{sid-ali.hamideche, florent.chiaroni, mohamed.rahal\}@vedecom.fr}}%
\thanks{$^{2}$ L2S-CNRS - CentraleSupelec - Univ Paris-Sud - Univ Paris-Saclay, 3 rue Joliot Curie, 91190 Gif-sur-Yvette, France
        {\tt\small Florent.Chiaroni@centralesupelec.fr}}%
}

\begin{document}

\maketitle
\thispagestyle{empty}
\pagestyle{empty}

%%%%%%%%%%%%%%%%%%%%%%%%%%%%%%%%%%%%%%%%%%%%%%%%%%%%%%%%%%%%%%%%%%%%%%%%%%%%%%%%
\begin{abstract}

Nowadays, autonomous driving systems can detect, segment, and classify the surrounding obstacles using a monocular camera. However, state-of-the-art methods solving these tasks generally perform a fully supervised learning process and require a large amount of training labeled data. On another note, some self-supervised learning approaches can deal with detection and segmentation of dynamic obstacles using the temporal information available in video sequences. In this work, we propose to classify the detected obstacles depending on their motion pattern. We present a novel self-supervised framework consisting of learning offline clusters from temporal patch sequences and considering these clusters as labeled sets to train a real-time image classifier. The presented model outperforms state-of-the-art unsupervised image classification methods on large-scale diverse driving video dataset BDD100K.

\end{abstract}

%%%%%%%%%%%%%%%%%%%%%%%%%%%%%%%%%%%%%%%%%%%%%%%%%%%%%%%%%%%%%%%%%%%%%%%%%%%%%%%%
\section{Introduction}

\label{sec:intro}

Building a vehicle with the capacity of driving itself is one of the most interesting and challenging applications of artificial intelligence. An autonomous vehicle must be aware of its environment, like identifying the obstacles, and should be able to make the right decisions to do the appropriate actions. These abilities can be developed independently and then merged to have a total or partial autonomy. Our work falls within the category of environment understanding.

This environment awareness property is acquired through the perception process, which enables interpreting the data provided by different kind of sensors, such as cameras, ultrasonic sensors, and LIDAR. For example the data can be the image of the surrounding obstacles and the perception process can be ensured by an object detection algorithm.

%\subsection*{Object detection}
\textbf{Object detection} \cite{DBLP:journals/corr/abs-1807-05511} consists of localizing and classifying objects in images or videos. It is an important feature in autonomous vehicles as it enables for instance to detect obstacles such as pedestrian, cars, or buildings.
During the last decade, deep learning approaches brought the tools for state-of-the-art detection methods. However, the best models in terms of prediction performances are fully supervised, such that they require a large number of annotated data. 
%\cite{5206848} \cite{Everingham15} \cite{journals/corr/LinMBHPRDZ14}
It turns out that achieving manually the labeling has a cost.
In order to reduce this limitation, other types of learning techniques have been investigated, like semi-supervised \cite{semi_sup} and weakly-supervised \cite{10.1093/nsr/nwx106} methods, as they can deal with partially unlabeled or noisy labeled data. In general, these techniques do not perform as well as supervised methods, but they are more practical by reducing the need of fully hand-labeled training sets. However, they still need some annotations. A more challenging task is to directly learn on easily acquirable unlabeled data.
%

%\subsection*{Unsupervised and self-supervised learning}
\textbf{Unsupervised learning} is the process of finding patterns using unlabeled data. More specifically, clustering methods \cite{clusterTech} are unsupervised learning approaches which consist of splitting data examples into groups depending on their similarities. These clusters can represent the semantics of objects, like cars and pedestrians.
Another category that does not use manually labeled data is \textbf{self-supervised learning} \cite{kolesnikov2019revisiting}. Unlike unsupervised methods, this still uses some kind of supervision which could be provided by other sensors or inferred from the data itself. A relevant example could be to learn to predict next frames in a video. This would not be possible if we do not know that frames in videos are temporally ordered.

\medskip

\begin{figure*}[t]
    \centering
    \includegraphics[width=\textwidth]{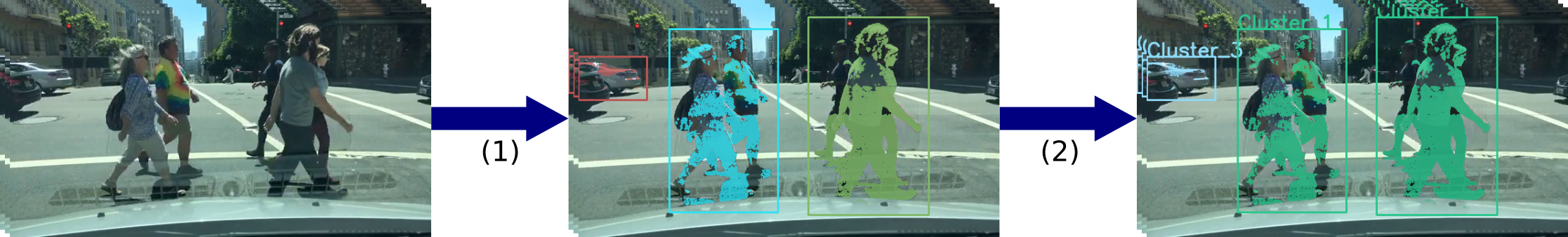}
    \caption{Overview of the proposed system inputs and outputs. (1) Dynamic obstacles detection, segmentation and tracking from a monocular camera video. (2) Unsupervised classification of dynamic obstacles.}
    \label{fig:overview}
\end{figure*}

%\subsection*{Overview}
In this work, we present a new self-supervised method for obstacles classification. The main contributions are to:
\begin{itemize}
    \item Exploit motion patterns as prior knowledge trough a state-of-the-art deep clustering method, for dynamic obstacles classification. The proposed approach outperforms state-of-the-art unsupervised image classification methods.
     \item Propose a temporally self-supervised framework, not using hand-labeled training data, for detection, segmentation and classification of dynamic obstacles.  
\end{itemize}
To the best of our knowledge, this is the first work that uses the temporal information provided by videos for unsupervised image classification of moving obstacles.
%
%\medskip
%
This work consists of two main parts. In the first part, we extract the patches from the videos while keeping their sequentiality. In the second part, we train a model to cluster these sequences and use the clusters as pseudo-labels to train an image classifier (Fig. \ref{fig:overview}). Since we prefer to have a variety of objects during short sequences rather than few objects during long sequences, the videos we use are captured by a static camera mounted on a stationary vehicle.

\medskip

The article is organized as follows. Sec. \ref{sec:related_work} presents the related work. Sec. \ref{sec:method} discusses motivations of this work and describes the proposed framework. Sec. \ref{sec:expermientation} presents and analyzes the comparative empirical experiments. Finally, Sec. \ref{sec:conclusion} presents conclusions and perspectives.

\section{Related work}
\label{sec:related_work}

\paragraph*{\textbf{Dynamic obstacle detection}}
Obstacle detection is an important feature for autonomous vehicles. Different works have tackled this problem especially for detecting dynamic obstacles using a monocular camera. While state-of-the-art methods are supervised  \cite{DBLP:journals/corr/abs-1807-05511}, some others instead proposed self-supervised approaches. In \cite{dyn_seg}, Guizilini \textit{et al.} focused on detecting all the pixels associated with dynamic obstacles. They start by computing the sparse optical flow using SIFT descriptor \cite{SIFT} to match the current frame with the previous one. Then, they use RANSAC algorithm \cite{Fischler:1981:RSC:358669.358692} to estimate the camera motion and consequently split the interest points into 2 sets representing respectively static and dynamic obstacles. Finally, they use these 2 sets to incrementally train a Gaussian Process (GP) classifier \cite{GP} to extrapolate the previous splitting stage for all the pixels of the image. As an extension to this method, Bewley \textit{et al.} \cite{quteprints69800} proposed to separate the detected dynamic bounding boxes into multiple instances.  This method adds the density based clustering algorithm DBSCAN \cite{Ester:1996:DAD:3001460.3001507}, to split the set of the dynamic interest points into multiple sets. Next, it uses a k-NN \cite{knn} trained on these sparse sets of pixels, in order to enable a dense prediction over all image pixels.

\paragraph*{\textbf{Image clustering}}
Data clustering is an active research topic in machine learning, especially for choosing the distance function and the appropriate evaluation metrics. Many popular clustering algorithms have emerged like K-means \cite{KMEANS}, GMM \cite{GMM} and spectral clustering \cite{SpecClus}. Combined with an upstream Principal Component Analysis (PCA) \cite{PCA}, these methods become interesting for a wider range of problems. But we cannot expect relevant semantic classification by applying them directly on image rows of pixels.

Some works, inspired by the recent performances of supervised methods and motivated by the feature learning capacity of deep neural networks, use auto-encoder \cite{AE} and its variants such as \cite{SWWAE} to learn latent representations. 
More recent works propose to combine the visual representation learning with the clustering in a single model. These methods generally work better as they jointly optimize both objectives. DEC \cite{DEC} learns simultaneously the cluster centers and the parameters of a deep neural network. To avoid degenerate solution the model requires a pre-training. DAC \cite{DAC} recasts the clustering problem into a binary pairwise-classification where the goal is to predict the same class for similar objects and different classes for dissimilar ones. DeepCluster \cite{DeepCluster} iteratively performs the following successive steps until convergence. First, it encodes the input images with deep convolutional feature extractors. Second, it associates the obtained encoded latent feature vectors with different cluster labels by using K-means. Finally, it trains the convolutional model to predict from input images these cluster labels.
IIC \cite{IIC} trains a model to maximize the mutual information between the representations of a paired dataset. The paired dataset is generated from original training images by using data augmentation techniques such as affine transformations. To the best of our knowledge, IIC presents the best prediction performances in the image clustering literature. Besides, IIC paired dataset strategy may enable to associate patches corresponding to a given instance observable on successive frames.

Recently, self-supervised representation learning appeared to solve pretext task.
These pretext tasks are often chosen to be as similar as possible to the original task.
An example of such tasks consists of predicting the context \cite{doersch2015unsupervised}, colors from gray-scale images \cite{DBLP:journals/corr/ZhangIE16}, or rotations \cite{gidaris2018unsupervised}. Some other strategies exploit the temporal signal, like minimizing the cosine distances between the representations of the same patch from different frames \cite{doersch2015unsupervised}, or predicting the motion segmentation from static images \cite{pathakCVPR17learning}. The learned representations empirically demonstrated the corresponding feature extractors usefulness.

Next section presents the proposed approach consisting of adapting the state-of-the-art image deep clustering algorithm IIC \cite{IIC} to a temporally self-supervised framework designed for dynamic obstacles analysis.

\section{Method}
\label{sec:method}
In this section, we present the proposed approach. We start by introducing the motivation and then we detail its functioning. 
\subsection{Motivation}
We propose to add the temporal information provided by videos to improve the performance of existing unsupervised classification methods as we apply them on detected dynamic obstacles. It turns out that directly exploiting visual features from independent images, without using the rest of the potentially associated temporal video sequence, is limiting for the following reasons:
\begin{itemize}
  \item Objects may have different shapes and colors depending on the viewpoint and the posture. The front-view of a car does not present the same shape as its side-view.
 \item We can see different parts of a same object, e.g. a vehicle with a trailer, as independent objects until we see it moving.
 \item Objects from different semantic classes may have similar features, like shape and texture, but they can be distinguishable by their way of moving; e.g., persons and cyclists are similar in shape while they move differently. 
\end{itemize}
Using video can help the algorithm to learn an abstract representation of each class that is independent from the viewpoint. For instance, an unsupervised algorithm trained on static data can assign the same label to different objects that are similar when observed from the same side (e.g., the front-view of a car and a truck). It can also assign different labels to the same object observed from different sides, such as a car observed from its front-view and its side-view. Moreover, temporal sequences catch the way the object moves. This may help the model to distinguish objects from different potentially moving obstacle categories, depending on their motion patterns. For instance, a pedestrian moves differently from a car. They are respectively deformable and solid objects.

\bigskip

The proposed self-supervised learning system is composed of 2 consecutive stages: During the former it extracts the dynamic obstacles from the videos. During the latter it trains a model to predict the class of the previously extracted obstacles. See Fig. \ref{fig:overview}.

\subsection{Dynamic obstacles extraction}
The goal in this part is to successively detect, track and extract the patches containing the dynamic objects and group them into sequences containing the same instance. We adapt to our problem the approach previously proposed by Bewley \textit{et al.}, 2014 \cite{quteprints69800}. Consequently, we design our dynamic obstacles extraction module as follows:

\paragraph{Sparse optical flow computing} In order to detect the moving obstacles, the first step is to compute the optical flow by matching the interest points of the current and the previous frame using a paralyzed version of SURF \cite{opencv_library}.

\paragraph{Static and dynamic obstacles separation} The goal here is to split the set of interest points, computed in the previous step, into 2 sets, one containing the static points and the other containing the dynamic ones. As we deal with videos obtained by a static camera (absence of camera movement), the partitioning in this step is straightforward. The ideal optical flow must be $\mathbf{0}$ for static points and must be different from $\mathbf{0}$ for dynamic points. However, in practice some matching noise can deteriorate the process. Consequently, we propose to use an empirical threshold close to zero to get static points. We also use a second higher arbitrary threshold to get only confident dynamic points. Furthermore, as we work from a static camera recording point of view, we do not need the RANSAC algorithm \cite{Fischler:1981:RSC:358669.358692} presented in the original method \cite{quteprints69800}.

\paragraph{Dynamic points distinction} In the previous step we acquired the dynamic points of interest that indicate the location of the moving obstacles. However, to be used for classification, we gather them by dynamic obstacle instances. In this part we follow the original method \cite{quteprints69800}, which uses DBSCAN \cite{Ester:1996:DAD:3001460.3001507}, to split the dynamic set into multiple sets corresponding to different obstacles. The main advantage of this method is that it does not require the number of clusters to be defined. The algorithm directly takes as inputs the set of the dynamic points represented by their position and optical flow velocity.

\paragraph{Obstacles tracking} To associate patches of obstacles observable from consecutive frames we use the tracking algorithm \textit{Simple Online and real-time Tracking} (SORT) \cite{8296962}. It is robust in terms of multiple object tracking accuracy and execution time.

\paragraph{Obstacles Segmentation} For the segmentation we apply a GMM-based background subtraction method \cite{MOG2} on consecutive frames and then mask the results with bounding boxes containing the dynamic obstacles.

\paragraph{Patch grouping} 
The last step is to extract, from all images, the square patches corresponding to bounding boxes of the tracked dynamic obstacles, and then group these patches into sets. Each set contains the temporally sorted patches of a given tracked instance.

\begin{figure}
    \centering
    \includegraphics[width=0.48\textwidth]{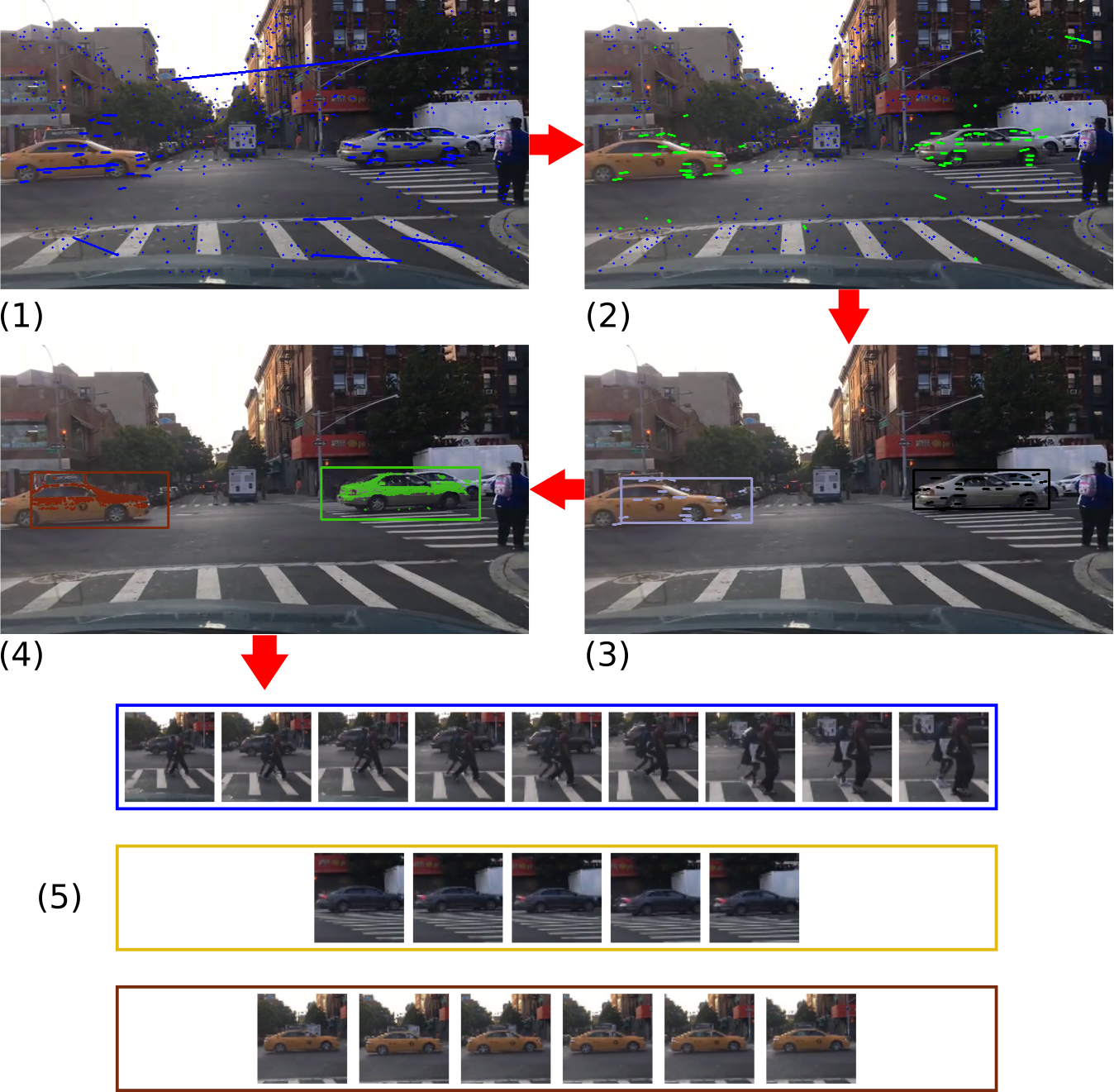}
    \caption{Dynamic obstacles extraction. (1) Optical flow estimation. (2) Detection of dynamic points of interest. (3) Dynamic obstacle instances detection. (4) Tracking and segmentation. (5) Patches extraction and grouping.}
    \label{fig:extraction}
\end{figure}

We sum up these steps in Fig. \ref{fig:extraction}. After having extracted moving obstacles, the second stage consists of categorizing them depending on their motion patterns.

\subsection{Obstacles classification}
In this section we explain the classification approach which consists of using unsupervised learning techniques to cluster the sequences of tracked obstacles, and then use the obtained clusters as pseudo-labels to train a classifier on single patches. This enables an immediate prediction using single patches instead of temporal sequences.
\begin{figure}
    \centering
    \includegraphics[width=0.486\textwidth]{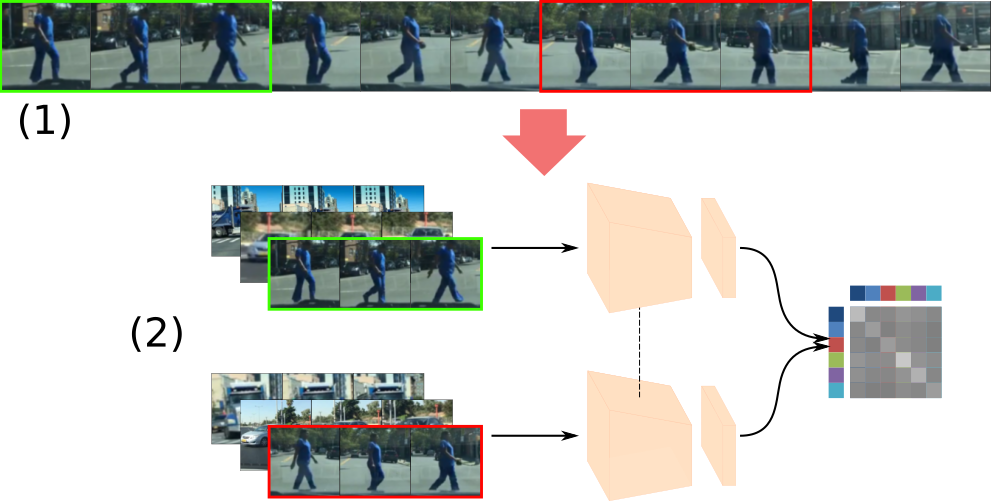}
    \caption{Clustering training using temporal sequences of instances. (1) Randomly sampling two sub-sequences from a given whole sequence to generate pairs. (2) IIC model \cite{IIC} training.}
    \label{fig:IIC}
\end{figure}

\subsubsection{Clustering of patch sequences}
In an unsupervised setting, clustering of real-world images is a challenging task. 
In this part, we explain how to exploit as a prior knowledge the temporal information concerning the instances to classify, in order to adapt the state-of-the-art clustering method IIC \cite{IIC} to our real-word application case.

%\textbf{Training:}
In order to train IIC \cite{IIC} for clustering, we use patch sequences extracted during the first part of our approach instead of single patches. As IIC needs a paired dataset, we generate pairs using two patch sub-sequences extracted from the same patch sequence, as illustrated in Fig. \ref{fig:IIC}. This gives to the model a prior on obstacle movements, and enables to associate patches corresponding to the same obstacle observed with varying postures and from different view-points.

As it is trained on sub-sequences, this deep clustering model cannot predict the classes for single patches. Consequently, in the next section we use cluster labels as pseudo-labels to train an image classifier over the extracted patches.

\subsubsection{Images classification}
\begin{figure*}
    \centering
    \includegraphics[width=0.6\textwidth]{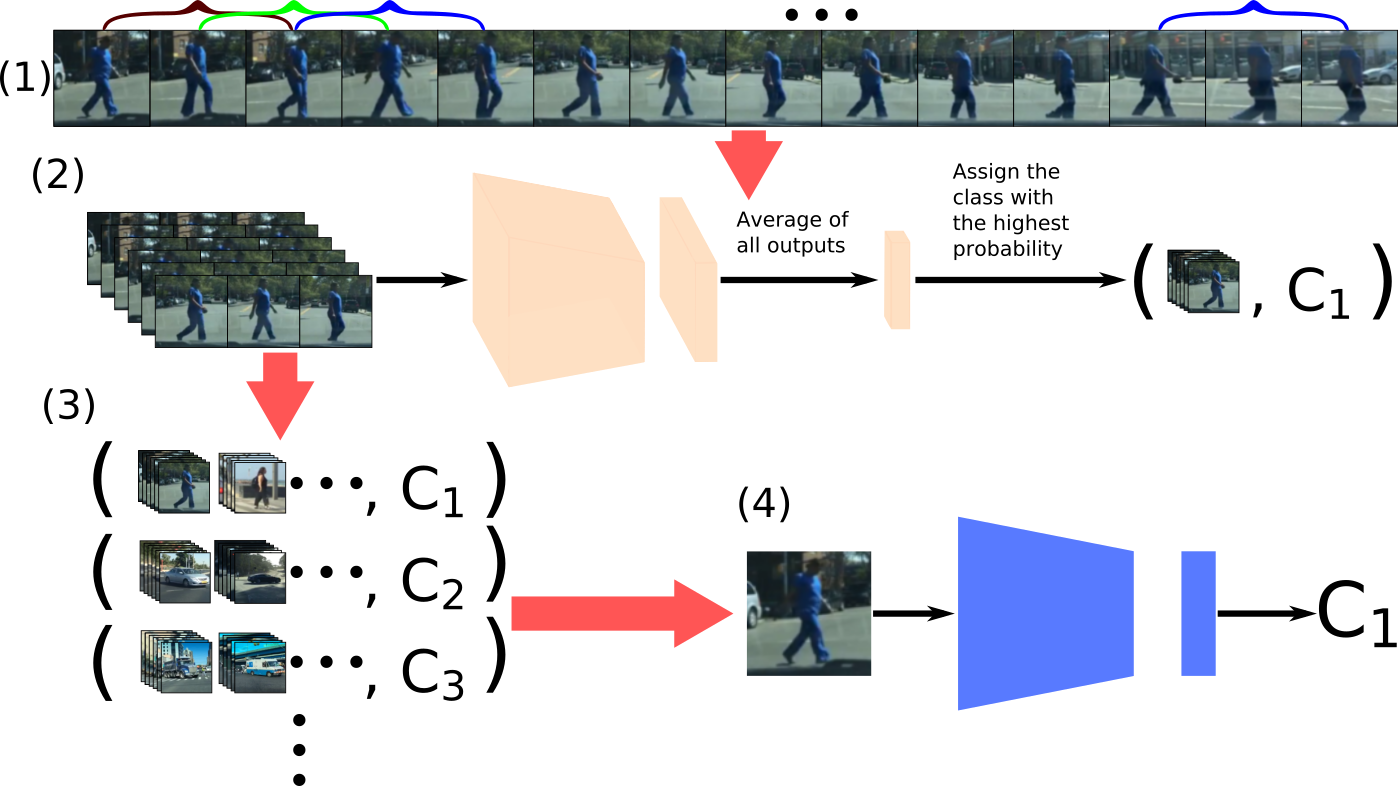}
    \caption{Image classification training process. (1) Selection of all the possible sub-sequences for a given sequence. (2) Trained IIC model \cite{IIC} predictions for all sub-sequences, and average estimation of these predictions in order to assign a cluster to all the patches contained in the corresponding whole sequence. (3) Creation of the classifier training dataset using the computed pseudo-labels corresponding to the previously identified clusters. (4) Image (i.e. patch) classifier training.}
    \label{fig:train}
\end{figure*}

As explained previously we trained IIC \cite{IIC} on sub-sequences to generate pseudo-labels. For each sequence of patches corresponding to the same obstacle $i$, as shown in Fig. \ref{fig:train}, we average over the softmax outputs of all possible sub-sequences $s_{ij}$. Then, we assign the class with the highest probability to all patches contained in the same sequence.
As a consequence, we can train an image classification model in a supervised manner on the obtained training set of patches, which are respectively associated with an independent pseudo-label. The whole process of this part is shown in Fig. \ref{fig:train}.

\section{Experiments}
\label{sec:expermientation}
In this section, we start by describing selected datasets and the learning system settings. Then we present some empirical experiments and the corresponding qualitative and quantitative results.
\subsection{Datasets}
\subsubsection{Training data}
As we work with videos captured by a static camera on a vehicle, we used a dataset satisfying this specific requirement. BDD100K \cite{DBLP:journals/corr/abs-1805-04687} contains 100,000 40-second videos in 30 frames-per-second which makes a total of 120,000,000 images. The dataset also provides GPS/IMU which we used to extract the frames when the vehicle is stationary. As the task is already difficult, we chosen, for a first step, to use only videos with clear weather and which are recorded in the daytime. The total number of exploited frames is 100109.

\subsubsection{Test data}
\begin{figure}
    \centering
    \includegraphics[width=0.486\textwidth]{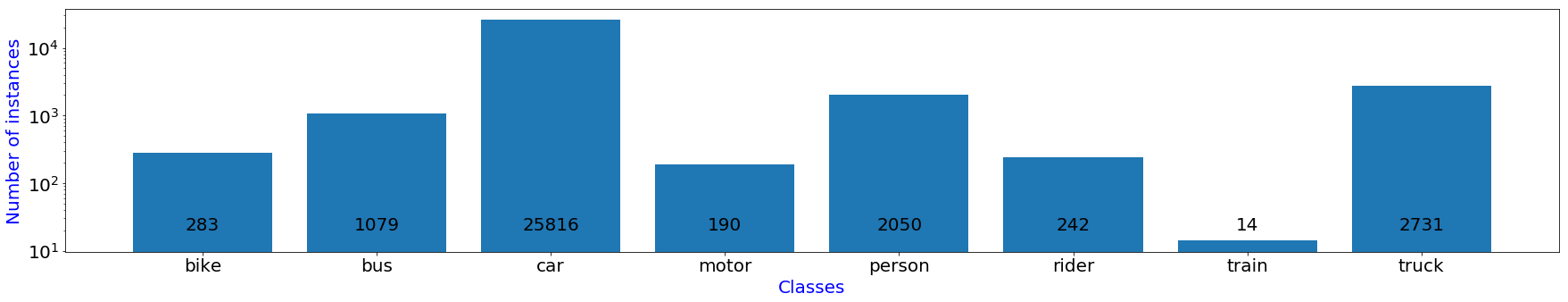}
    \caption{Distribution of classes in the generated dataset. In order the have a better visualization, the numbers of instances in the histogram are log-scaled.}
    \label{fig:classes_distribution}
\end{figure}

The BDD100k dataset \cite{DBLP:journals/corr/abs-1805-04687} also provides bounding box annotations of the frame at the $10^{th}$ second of each video. In order to evaluate our method, we use these bounding boxes to build a new dataset of images containing only one object similarly to the first part of our method, but here, we instead use the ground truth bounding boxes. We exclusively extracted potentially moving obstacles of the 8 classes \textit{Bicycle, Bus, Car, Motorcycle, Person, Rider, Train, Truck}. 
The classes distribution of the new built dataset is shown in Fig. \ref{fig:classes_distribution}.

\subsection{Patches extraction}
Because of light changes and other noises impacting the images matching, the dynamic obstacle detection method may produce a lot of false positives, such that some bounding boxes do not contain any dynamic obstacle. However, many of them can be filtered out by only keeping the boxes that are persistent through the frames. Consequently, we decide to keep patches that appear at least in 3 consecutive frames. We also used the SORT \cite{8296962} tracker to detect and reject the colliding objects. It turns out that sequences with length of 3 frames enable to capture dynamic obstacles movement patterns, while preserving enough data to train our model. The total number of patch sequences extracted is $3607$, with an average of $4.84$ images per instance. On another note, if we do the assumption that videos follow the same classes distribution than the created test dataset, then the distribution of bounding box images (i.e. patches) automatically extracted is probably considerably unbalanced. For instance, the \textit{car} class may be more represented than other classes.

\subsection{Inputs of the deep neural network models}
To train the first model which is the one used to generate pseudo labels, we use input sub-sequences of 3 images. We then apply the same random transformation to the patches of the hole sub-sequences. We use the data augmentation techniques proposed in the original implementation of IIC \cite{IIC}. This includes the random cropping and resizing to get fixed size $64 \times 64$.  We also apply Sobel filter as proposed in IIC \cite{IIC} and in DeepCluster \cite{DeepCluster}. This helps to prevent the model from clustering based on trivial cues such as color. This encourages instead to use more meaningful cues like shape. The input of the model is $6 \times 64 \times 64$, where $6$ represents the Sobel filter applied to approximate the $2$ derivatives along the horizontal and vertical pixel axis of the $3$ images. 

For the classifier, we use the original patches as inputs. In other words, the input size is $3 \times 64 \times 64$, with $3$ representing the color channels number. We also apply data augmentation techniques such as random horizontal flip and cropping.

\subsection{Training}
We use the same configuration as IIC \cite{IIC} to train our sequence clustering model. We use the Adam optimizer \cite{Adam} with a learning rate $10^{-4}$, and the objective functions are optimized alternatively. For the classifier, we use the deep convolutional architecture Resnet-34 \cite{he2016deep} with Adam optimizer with a learning rate $10^{-3}$ and the cross-entropy loss function.

\subsection{Evaluation metrics}
In our experiment, similarly to other works on unsupervised images classification, we use the standard clustering accuracy (ACC) metric. It consists of finding the best one-to-one mapping between the ground truth labels and the clusters. In order to better evaluate our unbalanced dataset, we apply up-sampling before the evaluation, such that every cluster includes the same number of patches.

\subsection{Results analysis}
We compare the proposed approach exploiting the temporal information with the state-of-the-art image clustering method IIC \cite{IIC}. As the patches automatically extracted with the presented detection algorithm can be different from the ground truth, we assume that the clustering algorithm learns the same classes. We evaluated the prediction performances of our method with two settings. The former is a multi-class classification and the latter is a binary classification between car and person.

\subsubsection{Multi-class classification}
As shown in Table \ref{tab:multi_acc}, the proposed method outperforms the original IIC and DeepCluster.
 \begin{table}
     \centering
     \begin{tabular}{lc}
         \hline
          Method & ACC \\
          \hline
          PCA \cite{PCA} + K-means \cite{KMEANS} & $0.145$ \\
          PCA \cite{PCA} + GMM \cite{GMM} & $0.142$ \\
          DeepCluster \cite{DeepCluster} & $0.179$ \\
          IIC \cite{IIC} & $0.239$ \\
          Ours & $\boldsymbol{0.313}$ \\
          \hline
     \end{tabular}
     \caption{Balanced accuracy (ACC) scores for the multi-class classification.}
     \label{tab:multi_acc}
\end{table}

\subsubsection{Binary classification}
\begin{figure}
    \centering
    \begin{subfigure}[b]{0.18\textwidth}
        \centering
        \includegraphics[width=1\textwidth]{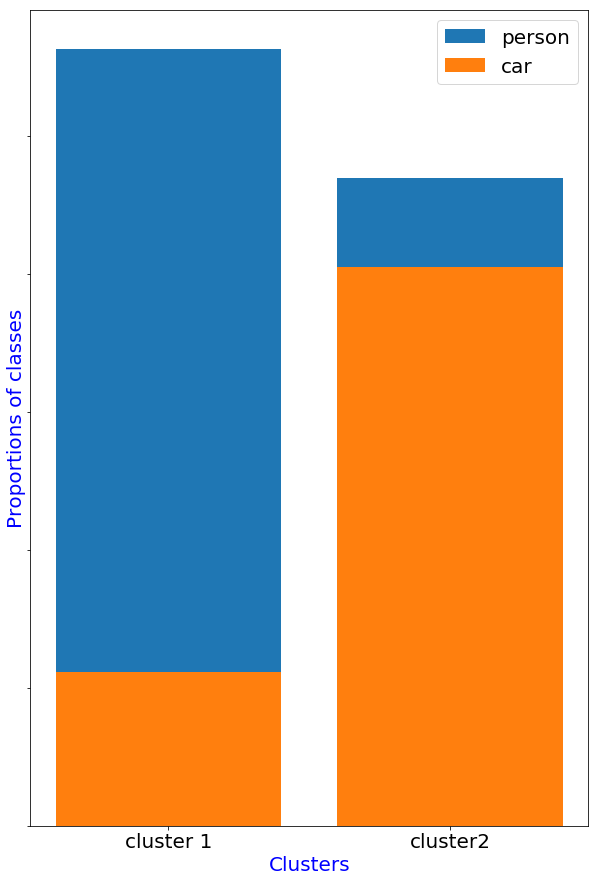}
        \caption{Our method}
    \end{subfigure}%
    ~
    \begin{subfigure}[b]{0.18\textwidth}
        \centering
        \includegraphics[width=1\textwidth]{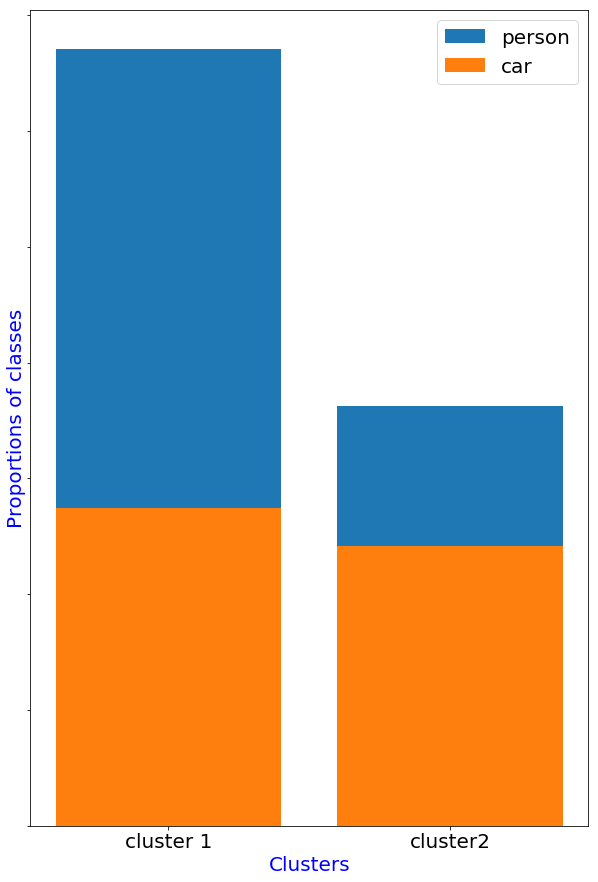}
        \caption{Original IIC \cite{IIC}}
    \end{subfigure}
    \caption{Proportions of classes over clusters for the binary Classification. Our method predicts more homogeneous clusters than the original IIC \cite{IIC}.}
    \label{fig:histo_binary}
\end{figure}

\begin{figure}
    \centering
    \begin{subfigure}[b]{0.46\textwidth}
        \centering
        \includegraphics[width=1\textwidth]{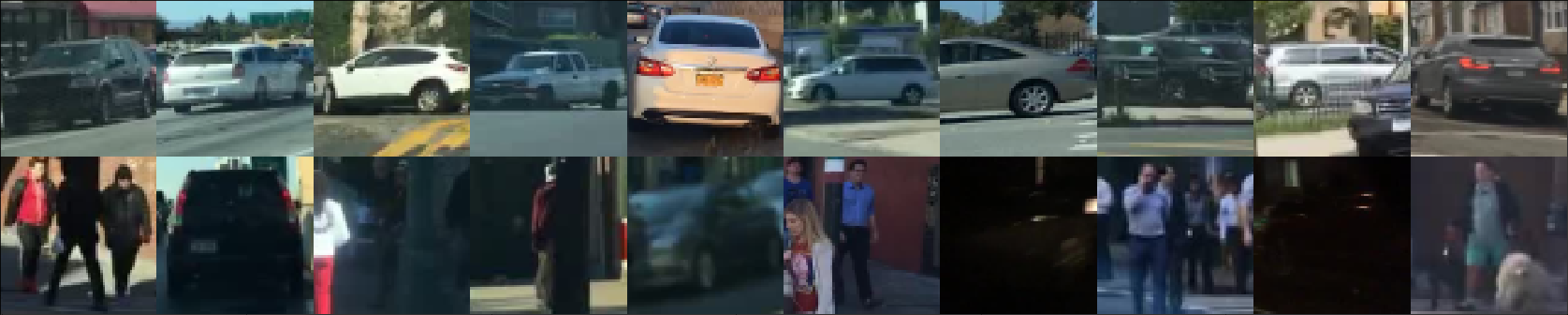}
        \caption{Our method}
    \end{subfigure}%
    \\
    \begin{subfigure}[b]{0.46\textwidth}
        \centering
        \includegraphics[width=1\textwidth]{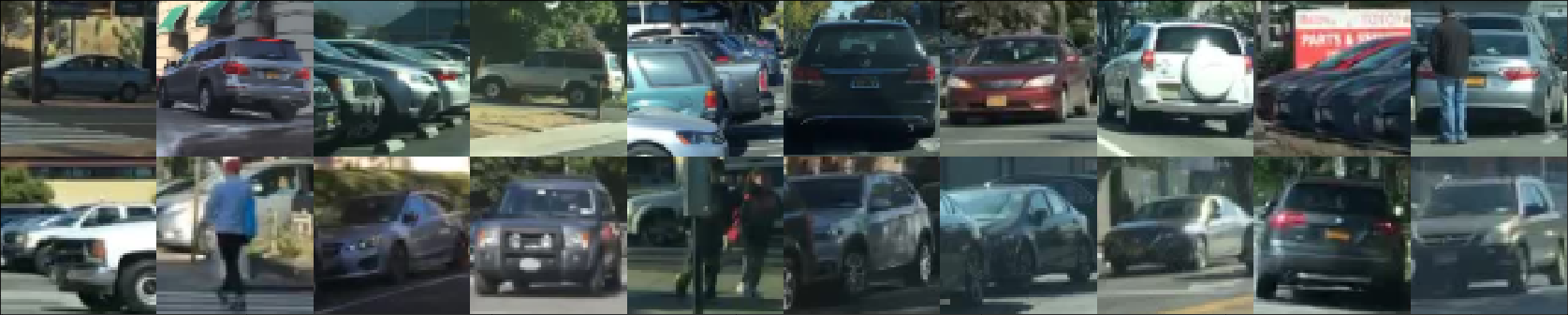}
        \caption{Original IIC \cite{IIC}}
    \end{subfigure}
    \caption{Moving obstacles with the highest probabilities to be in their respective cluster, such that each row corresponds to a distinct cluster.}
    \label{fig:cluster_2_classes}
\end{figure}

Table \ref{tab:binary_acc} shows that the proposed method significantly outperforms the original IIC \cite{IIC} as well in binary classification: While IIC presents the accuracy score 0.569, relatively close to an average of random predictions, the proposed extension presents the accuracy score 0.829. We believe that this is due to the difference in terms of movement patterns between a car and person which is only observable by temporal information. The histograms in Fig. \ref{fig:histo_binary} also show that our method is able to detect the similarities between instances of the class person, as they are mainly concentrated in one cluster. The class car is divided due to the unbalanced composition of the training dataset BDD100K. Fig. \ref{fig:cluster_2_classes} shows obstacles patches with the highest probabilities to be in their associated cluster. In contrast to IIC, the most confident examples per-cluster of the proposed method respectively represent cars and pedestrians.
\begin{table}
    \centering
    \begin{tabular}{lc}
        \hline
        Method & ACC \\
         \hline
         IIC \cite{IIC} & $0.569$ \\
        %  \hline
         Ours & $\boldsymbol{0.829}$ \\
         \hline
    \end{tabular}
    \caption{Balanced accuracy (ACC) score comparison, with the original best state-of-the-art method IIC \cite{IIC}, for the binary classification \textit{car-versus-person}.}
    \label{tab:binary_acc}
\end{table}

\begin{table}[t]
\centering
% \resizebox{0.5\textwidth}{!}{
    \begin{tabular}{l|c|c|}
    \cline{2-3}
                                                                  & Proposed & Bewley \textit{et al.} \\
                                                                  & Method &  \cite{quteprints69800}\\
                                                                  \hline
    \multicolumn{1}{|l|}{Sparse optical flow computing}           & 0.03            & 0.35                  \\ 
    \multicolumn{1}{|l|}{Static and dynamic obstacles separation} & 0.01            & 0.03                  \\ 
    \multicolumn{1}{|l|}{Dynamic points distinction}              & 0.009           & 0.009                 \\
    \multicolumn{1}{|l|}{Obstacles tracking}                      & 0.001           & \multirow{2}{*}{2.03} \\
    \multicolumn{1}{|l|}{Obstacles Segmentation}                  & 0.001           &                       \\ 
    \multicolumn{1}{|l|}{\textbf{Obstacles Classification (Contribution)}}                & 0.07            & -                     \\ \hline
    \multicolumn{1}{|l|}{\textbf{Total}}                          & \textbf{0.12}   & \textbf{2.5}          \\ \hline
    \end{tabular}
    \caption{Computational time in seconds of our method compared to Bewley \textit{et al.}, 2014 \cite{quteprints69800}. The results presented in this table are from our implementation of both methods.}
    \label{tab:compute_cost}
% }
\end{table}

\begin{figure*}
    \centering
    \begin{subfigure}[b]{0.486\textwidth}
        \centering
        \includegraphics[width=1\textwidth]{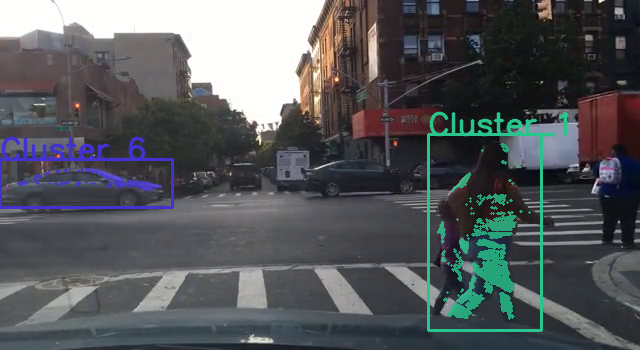}
        % \caption{Our method.}
    \end{subfigure}%
    ~
    \begin{subfigure}[b]{0.486\textwidth}
        \centering
        \includegraphics[width=1\textwidth]{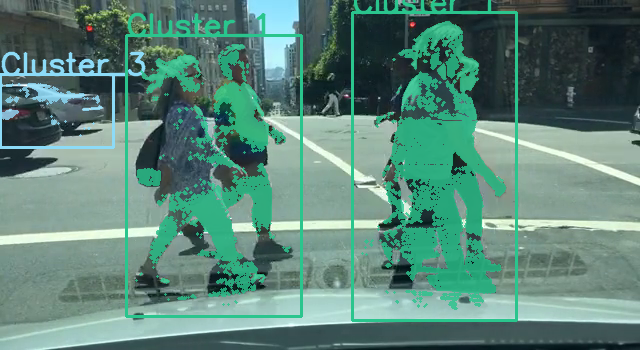}
        % \caption{Original IIC \cite{IIC}.}
    \end{subfigure}
    \caption{Examples of our method applied to videos for temporally self-supervised detection, segmentation, and \textbf{classification} of moving obstacles.}
    \label{fig:semantic_segmentation}
\end{figure*}

\subsection{Semantic instance segmentation}
To sum up on these experiments, Fig. \ref{fig:semantic_segmentation} illustrates some output predictions of the proposed complete framework when it is applied on videos from a static point of view. It performs detection, segmentation, and unsupervised classification of moving obstacles. We can observe that the \textit{Cluster 1} assigned to pedestrian bounding boxes is different to the clusters assigned to car vehicles. 

However, a weakness of the proposed framework is that the detection process does not enable to correctly separate pedestrian instances if they visually overlap with each other. An additional depth map information may help to deal with this issue while improving in the meantime the foreground dynamic obstacles segmentation.

\subsection{Computational cost}
The method was implemented with Python 3, OpenCV 4 \cite{opencv_library} and PyTorch \cite{PyTorch}. We did the experiments on a machine equipped with an Intel Core i-4710HQ CPU, a GTX 970M GPU and 16GB of RAM. In Table \ref{tab:compute_cost} we compare in terms of execution time our method to the method proposed by Bewley \textit{et al.} \cite{quteprints69800} as we took inspiration from the latter. We show that we can save 2 seconds by substituting tracking and segmentation steps. This is mainly due to the fact that that we substituted the time consuming k-NN part with a background subtraction algorithm \cite{MOG2} \cite{MOG2} as in our case we work with videos without ego-vehicle camera motion.

\section{Conclusions and perspectives}
\label{sec:conclusion}
To sum up, motivated by the drive to avoid hand labeled training data for the classification of moving obstacles, we have proposed in this article a novel unsupervised image classification approach. It exploits the temporal information concerning the visual pose transformations and motion patterns of the observed moving obstacles. In practice, we have integrated this technique in a self-supervised learning framework in order to jointly detect, segment, and classify the moving obstacles from a monocular camera without any pre-training or hand labeled data. Our empirical study on BDD100K dataset has demonstrated the usefulness and competitiveness of the proposed framework compared to the state-of-the-art techniques in terms of computational cost for the detection and segmentation steps, and in terms of prediction performances concerning the image classification task without hand labeled data. However, compared to fully supervised techniques using annotated data, the proposed model remains limited in terms of prediction performances. Thus, it may be interesting to investigate future research for:
\begin{itemize}
    \item Improving detection and tracking parts in order to provide more consistent patch sequences in input of the proposed temporal clustering approach;
    \item Using additional information from a depth sensor in order to better separate and analyze moving obstacles;
    \item Dealing with unbalanced training data for unsupervised classification without class proportion prior knowledge.
\end{itemize}

% \section*{ACKNOWLEDGMENT}

% The preferred spelling of the word `acknowledgment' in America is without an `e' after the `g'. Avoid the stilted expression, `One of us (R. B. G.) thanks . . .'  Instead, try `R. B. G. thanks'. Put sponsor acknowledgments in the unnumbered footnote on the first page.

%%%%%%%%%%%%%%%%%%%%%%%%%%%%%%%%%%%%%%%%%%%%%%%%%%%%%%%%%%%%%%%%%%%%%%%%%%%%%%%%

% References are important to the reader; therefore, each citation must be complete and correct. If at all possible, references should be commonly available publications.
% They are cited like so: \cite{IEEEexample:articleetal}, \cite{IEEEexample:book}, ...

\bibliographystyle{IEEEtran}
\bibliography{IEEEabrv,IEEEexample}

\end{document}